\documentclass[]{youtu}

\usepackage{times}
\usepackage{latexsym}

\usepackage[T1]{fontenc}

\usepackage[utf8]{inputenc}

\usepackage{microtype}

\usepackage{inconsolata}

\usepackage{graphicx}
\usepackage{subcaption}
\usepackage{algorithm}
\usepackage{algorithmic}
\usepackage{booktabs}
\usepackage{makecell}
\usepackage{bm}
\usepackage{pifont}
\usepackage{wrapfig}
\usepackage{hyperref}
\usepackage{amsmath}
\usepackage{amsthm}
\usepackage{amsfonts}
\usepackage{amssymb}
\usepackage{multirow} 
\usepackage{color}

\usepackage{newfloat}
\usepackage{listings}
\usepackage{enumitem} 

\newcommand{\sysname}{\textbf{\textit{ContextPilot}}}

\usepackage{booktabs}
\usepackage{xcolor}
\usepackage{colortbl}
\usepackage{multirow}
\usepackage{tabularx}

\usepackage[most]{tcolorbox}
\usepackage{fontawesome5}
\definecolor{yellow}{HTML}{F6BD60}
\usepackage{multicol}

\definecolor{lightgreen}{rgb}{0.55, 0.71, 0.0}
\definecolor{bisque}{rgb}{0.87, 0.72, 0.53}
\definecolor{lightyellow}{rgb}{0.99, 0.76, 0.0}
\definecolor{lightblue}{rgb}{0.36, 0.54, 0.66}
\definecolor{darkgray}{rgb}{0.66, 0.66, 0.66}
\definecolor{salmon}{rgb}{0.98, 0.50, 0.45}
\definecolor{deeppurple}{rgb}{0.4, 0.0, 0.4}

\definecolor{yellow}{HTML}{F6BD60}
\definecolor{white}{HTML}{FFE0C1}
\definecolor{pink}{HTML}{F5CAC3}
\definecolor{tale}{HTML}{84A59D}
\definecolor{red}{HTML}{F28080}
\definecolor{orange}{HTML}{FF7F00}
\definecolor{green1}{HTML}{72C3A3}
\definecolor{green2}{HTML}{70B48F}
\definecolor{orange}{HTML}{FE8019}
\definecolor{grey}{HTML}{EBDBB2}
\definecolor{brain}{HTML}{FFABBE}
\definecolor{blue}{HTML}{A3B7CA}
\definecolor{purple}{HTML}{5861AC}
\definecolor{narrative}{HTML}{458588}
\definecolor{white2}{HTML}{F8F5E9}
\definecolor{tablewhite}{HTML}{E5E1DA}
\definecolor{verylightgrey}{HTML}{CDCDCD}

\newcolumntype{P}[1]{>{\centering\arraybackslash}p{#1}}

\usepackage[switch]{lineno}

\usepackage{mathpazo}
\usepackage{graphicx}
\usepackage{natbib}
\title{ContextPilot: Teaching Agents for Proactive Context Management via Fine-grained RL}

\author{
  Zhuoshi Pan\textsuperscript{1,2 $\clubsuit$}, 
  Qizhi Pei\textsuperscript{3 $\clubsuit$},
  Junru Lu\textsuperscript{2},
  Honglin Lin\textsuperscript{3}, 
  H. Vicky Zhao\textsuperscript{1$~\heartsuit$},
  Di Yin\textsuperscript{2}, 
  Xing Sun\textsuperscript{2$~\heartsuit$}
}

\affiliation{\textsuperscript{1}Tsinghua University\quad\textsuperscript{2}Tencent Youtu Lab\quad\textsuperscript{3}Shanghai AI Lab}

\project{https://tencent.github.io/ContextPilot}
\sourcecode{https://github.com/Tencent/ContextPilot}
\model{https://huggingface.co/collections/panzs19/contextpilot}
\correspondence{$\clubsuit$~Equal Contribution; $\heartsuit$~Corresponding Authors.}

\begin{document}

\abstract{Long-horizon agentic tasks require large language models (LLMs) to iteratively retrieve, integrate, and maintain dispersed information across multi-turn interactions, but preserving all interaction histories leads to a continuously growing working context. Recent proactive context management methods allow models to edit their own working context with specialized tools, yet they still face \textbf{three key limitations}: (1) a \textit{limited toolset} restricted to search, deletion, and summarization, with no support for global planning, long-term memory, and adaptive compression; (2) \textit{inefficient exploration} that treats context management actions uniformly despite their heterogeneous impacts on final outcomes; and (3) \textit{coarse-grained credit assignment} that assigns the final trajectory-level reward to all intermediate context editing actions during RL. To bridge these gaps, we introduce \sysname{}, a proactive context management framework for long-horizon agentic reasoning. Our approach systematically augments the toolset with planning, long-term memory, and soft context offloading tools. We further propose an RL method tailored for context management, which uses context and entropy variation to identify critical editing decisions for branch sampling and estimates action-level advantages from all branched trajectories that pass through the corresponding context editing action. Experiments on long-context QA and deep search tasks show that \sysname{} achieves stronger performance with a more compact working context, consistently outperforming existing baselines across various base models and benchmarks.}

\maketitle

\section{Introduction}

\begin{figure*}[ht]
\centering
\includegraphics[width=\textwidth]{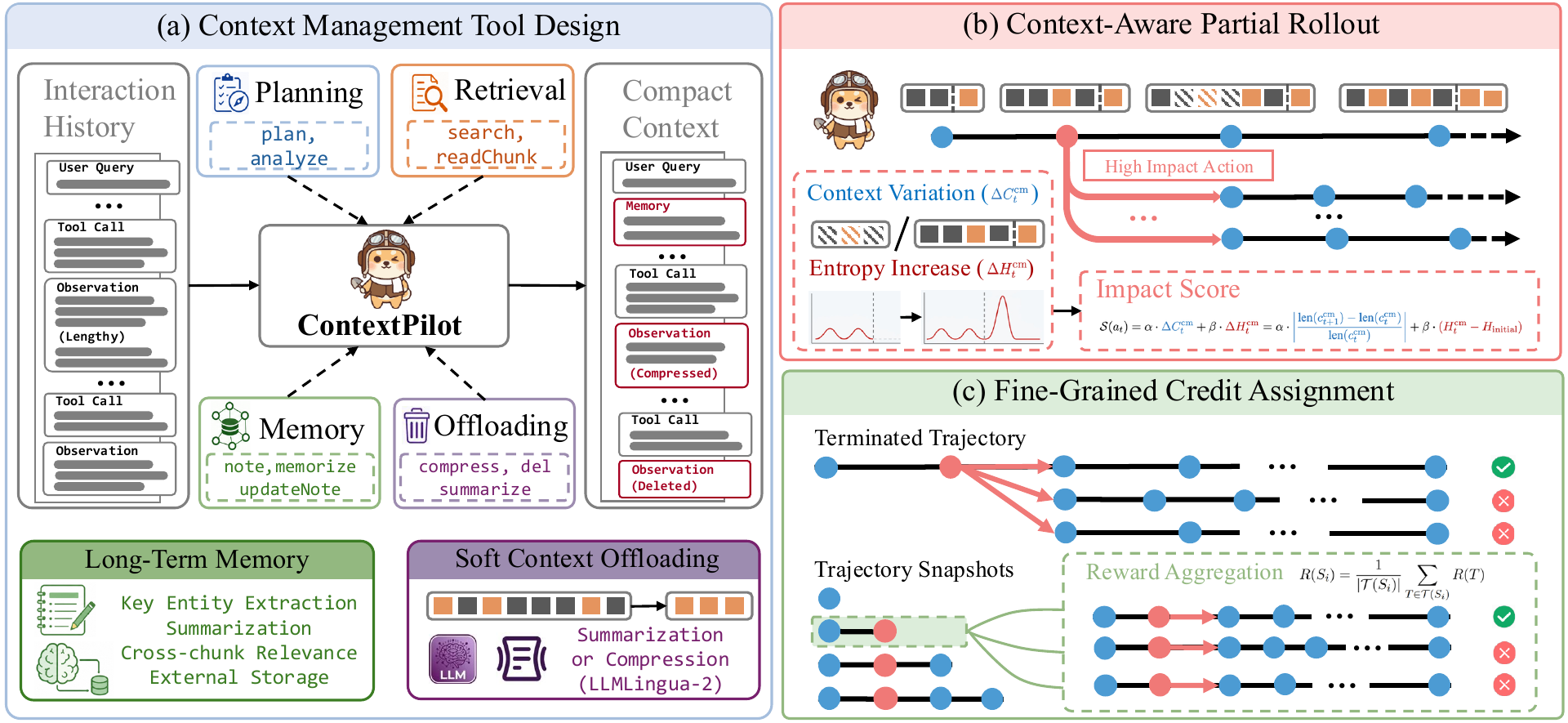}
\caption{
Overview of \sysname{}: (a) Extending the context management tools to support planning, long-term memory, and soft context offloading; (b) Improved RL training algorithm with context-aware partial rollout and (c) fine-grained snapshot-level credit assignment.
}
\label{fig:main_figure}
\end{figure*}

Large language models (LLMs) have shown strong reasoning capabilities on information-intensive tasks, including long-context QA~\cite{zhang2024infty,du2025simvbg} and deep search~\cite{bc_en,team2025tongyi}.
These tasks require models to identify key evidence from long contexts, integrate complex relations, and, when necessary, seek information through multi-turn tool interactions~\cite{feng2025retool}.
Prevailing agentic paradigms like ReAct~\cite{yao2023react} typically append prior reasoning, tool calls and tool responses to the context as the interaction proceeds, resulting in a rapidly growing context.
Prior studies~\cite{shao2025foldact,wu2025resum,lu2025scaling,su2026u} address context growth through \textit{human-designed workflows}, where truncation or summarization is triggered by fixed rules.
Although straightforward, they give the model no control over its own context, and are difficult to adapt to different scenarios~\cite{liu2026pensieve}.
To alleviate this, recent works~\cite{li2025sculptor,ye2025agentfold,liu2026pensieve} propose \textbf{\textit{proactive context management}}, allowing models to spontaneously manage their working context through context editing tools.
Despite its flexibility, this paradigm still faces three key limitations.

\textbf{First}, existing context management toolsets, which typically consist of search, deletion, and summarization, are insufficient for managing the context of long-horizon tasks.
Effective context management requires more than simply removing or summarizing content: models also need to build long-term memory across scattered fragments~\cite{chhikara2025mem0,xu2026mem}, maintain structured entity-event episodes~\cite{lu2026structured}, and plan globally before taking subsequent actions~\cite{yao2023react,liu2026plan}.
\textbf{Second}, although recent studies~\cite{yu2026agentic,liu2026pensieve} have applied RL fine-tuning to help models further familiarize with context editing tools, \textit{their training procedures are not specifically adapted for context management}.
Unlike common tools, context editing tools can substantially overwrite the interaction history and exert a larger influence on subsequent steps~\cite{shao2025foldact}.
As shown in Figure~\ref{fig:pilot_study}, trajectory branches originating from different context management operations exhibit substantially different variance in their final success rates, indicating that certain operations are more sensitive than others.
Unfortunately, traditional RL relies on trajectory-level rollouts~\cite{dong2025agentic}, limiting adaptive exploration for various context management decisions.
\textbf{Third}, existing training procedures directly assign the final trajectory-level reward to all intermediate context management actions, neglecting fine-grained credit assignment~\cite{li2026salt}.

To address these challenges, we propose \sysname{}, a proactive context management framework for long-horizon agentic reasoning.
Our contributions are threefold:
(1) On the \textbf{tool} side, our framework extends the existing \textbf{\textit{context management toolset}} with planning, long-term memory, and soft context offloading tools, enabling the agent to better control its context across long-horizon interactions.
(2) On the \textbf{training} side, we design an RL training method tailored to context management.
We propose \textbf{\textit{context-aware partial rollout}}, which uses context and entropy variation to identify critical context management actions for branch sampling.
For credit assignment, we incorporate rewards from all subsequent branches to estimate the advantage of an intermediate context management action, yielding a more \textbf{\textit{fine-grained reward signal}}.
(3) On the \textbf{evaluation} side, whereas prior work mostly focuses on long-context QA, we extend the evaluation to deep search tasks. Experiments on both tasks show that \sysname{} maintains a more compact context while achieving superior performance, bringing consistent gains and surpassing existing baselines.

\section{Related Work}

\subsection{Passive Context Management}
To handle growing interaction histories, existing methods typically manage historical context through predefined rules, such as truncating or summarizing messages once the context exceeds a length threshold.
Some methods~\cite{lu2023memochat,pan2025secom,wu2025resum,su2026u} are training-free and rely on fixed workflows or external modules to summarize, compress, or extract key information from historical context.
Another line of work~\cite{lu2025scaling,shao2025foldact} incorporates context folding into RL training, enabling models to summarize context and resume reasoning from the compressed state.
Although these methods reduce context length, the model remains a passive recipient of rule-managed context rather than an active manager that can decide \emph{when} and \emph{how} to manage its working context.

\subsection{Agentic Proactive Context Management}
Beyond the passive paradigm, recent work equips agents with context management tools, allowing agents to actively decide \emph{when} and \emph{how} to edit their working context.
MemGPT~\cite{packer2023memgpt} is an early framework that treats an agent's memory as virtual memory and manages it through tool calls.
\citet{yu2026agentic} and \citet{xu2026mem} propose memory editing tools, training the agent to control its memory without an external controller.
Sculptor~\cite{li2025sculptor}, StateLM~\cite{liu2026pensieve}, and MemAct~\cite{zhang2025memory} provide models with context editing tools 
such as fragmentation, search, summarization, and deletion, and fine-tune models to use these tools.
AgentFold~\cite{ye2025agentfold} uses SFT to learn multi-scale folding operations that can condense the history. 
Despite demonstrating the promise of proactive context management, existing methods are still constrained by a narrow set of context management tools, and their training procedures fail to consider the varying impact of these tools.
In contrast, \sysname{} improves proactive context management by enriching tool design and refining training paradigms.

\section{Preliminary}
\label{sec:preliminary}

Before introducing \sysname{}, we first revisit the concepts of context management in agentic reasoning and conduct a pilot study.

\subsection{From Passive to Proactive: Context Management in Agentic Reasoning}
\label{subsec:proactive_context}
In prevailing agentic reasoning paradigms like ReAct~\cite{yao2023react}, 
the agent can perform reasoning and invoke tools to interact with an environment $\mathcal{E}$.
Formally, given the user query $c_0 = q$, at the $i$-th step, the LLM $\pi_\theta$ generates thought $t_i$ and tool call $a_i$ based on the current context $c_i$:
\begin{equation}
(t_i, a_i) \sim \pi_\theta(\cdot \mid c_i).
\end{equation}
In the next step, the generated thought $t_i$ and tool call $a_i$, together with the environment feedback $o_i \sim \mathcal{E}(\cdot \mid a_i)$, are appended to the context $c_{i+1} = c_i \oplus (t_i, a_i, o_i)$.
The model then iteratively continues this cycle until accomplishing the task.
As agentic reasoning proceeds, the context grows monotonically.
To mitigate the context overload, \textbf{\textit{proactive context management}} enables models to manipulate their context through specific tools, such as search, retrieval, deletion, and summarization. 
Specifically, let $\mathcal{A}$ denote the set of available actions, and $\mathcal{A}_{\mathrm{cm}} \subseteq \mathcal{A}$ denote the subset of context management actions.
Among them, we distinguish context editing actions as $\mathcal{A}_{\mathrm{ce}} \subseteq \mathcal{A}_{\mathrm{cm}}$, which directly modify the interaction history.
At step $i$, when the model invokes a context management action $a_i^{\mathrm{cm}} \in \mathcal{A}_{\mathrm{cm}}$, its context is updated by a context transition function $\mathcal{F}$:
\begin{equation}
c_{i+1} = \mathcal{F}(c_i, a_i^{\mathrm{cm}}) \oplus (t_i, a_i, o_i).
\end{equation}
Under this paradigm, training solely on the final trajectory is inadequate, as context editing operations (i.e., $a_i^{\mathrm{ce}}\in \mathcal{A}_{\mathrm{ce}}$) will modify historical messages.
To address this, existing studies~\cite{li2025sculptor,liu2026pensieve} adopt \emph{trajectory snapshots} and \emph{token-level loss masking}.
Suppose a trajectory $T$ contains $K$ tool invocations $\{a_i\}_{i=1}^{K}$, of which $M$ are context editing operations.
These context editing operations segment the multi-step trajectory into $M+1$ independent trajectory snapshots $\mathcal{S} = \{S_1, S_2, \dots, S_{M+1} \}$.
Each snapshot is then treated as an independent training instance, ensuring that every intermediate context state is incorporated in training.
Furthermore, token-level loss masking is applied to exclude the loss of outputs that have already appeared in previous trajectory snapshots.
This prevents redundant optimization over earlier tool calls and improves training stability and efficiency.

\begin{wrapfigure}{r}{0.5\columnwidth}
\centering
\includegraphics[width=\linewidth]{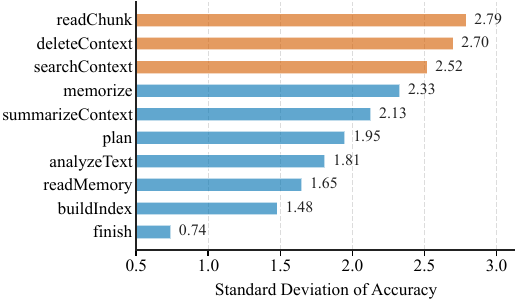}
\caption{
Impact of different tools on final outcomes: for each tool, we branch from it, sample $10$ continuation rollouts, and report the standard deviation of their task success rates.
Results are evaluated with Qwen3-8B on NovelQA~\cite{wangnovelqa}.
}
\label{fig:pilot_study}
\vspace{-10pt}
\end{wrapfigure}

\begin{figure}[t]
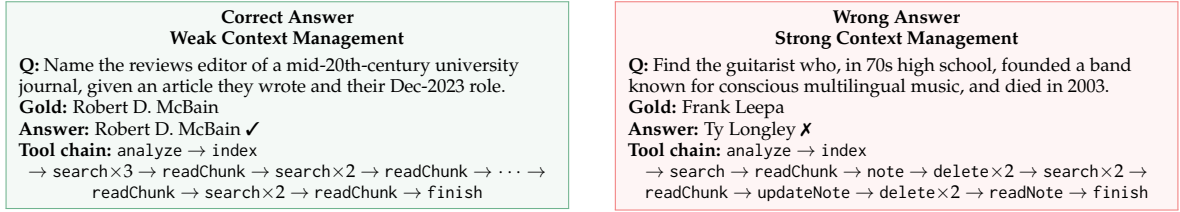

\centering
\scriptsize
\setlength{\fboxsep}{3pt}
\setlength{\tabcolsep}{2pt}
\renewcommand{\arraystretch}{1.08}
\begin{tabular}{@{}p{0.48\columnwidth}p{0.48\columnwidth}@{}}
\fcolorbox{green2}{green2!8}{
\begin{minipage}[t]{0.43\columnwidth}
\centering
\textbf{Correct Answer}\\
\textbf{Weak Context Management}

\vspace{2pt}
\raggedright
\textbf{Q:}
Name the reviews editor of a mid-20th-century university journal, given an article they wrote and their Dec-2023 role.

\textbf{Gold:}
Robert D. McBain

\textbf{Answer:}
Robert D. McBain \ding{51}

\noindent\textbf{Tool chain:} \texttt{analyze} $\rightarrow$ \texttt{index}\par
{\centering
$\rightarrow$ \texttt{search}$\times3$
$\rightarrow$ \texttt{readChunk}
$\rightarrow$ \texttt{search}$\times 2$
$\rightarrow$ \texttt{readChunk}
$\rightarrow \cdots \rightarrow$ \texttt{readChunk}
$\rightarrow$ \texttt{search}$\times2$
$\rightarrow$ \texttt{readChunk}
$\rightarrow$ \texttt{finish}
\par}
\end{minipage}
}
&
\fcolorbox{red}{red!8}{
\begin{minipage}[t]{0.43\columnwidth}
\centering
\textbf{Wrong Answer}\\
\textbf{Strong Context Management}

\vspace{2pt}
\raggedright
\textbf{Q:}
Find the guitarist who, in 70s high school, founded a band known for conscious multilingual music, and died in 2003.

\textbf{Gold:}
Frank Leepa

\textbf{Answer:}
Ty Longley \ding{55}

\noindent\textbf{Tool chain:} \texttt{analyze} $\rightarrow$ \texttt{index}\par
{\centering
$\rightarrow$ \texttt{search}
$\rightarrow$ \texttt{readChunk}
$\rightarrow$ \texttt{note}
$\rightarrow$ \texttt{delete}$\times2$
$\rightarrow$ \texttt{search}$\times2$
$\rightarrow$ \texttt{readChunk}
$\rightarrow$ \texttt{updateNote}
$\rightarrow$ \texttt{delete}$\times2$
$\rightarrow$ \texttt{readNote}
$\rightarrow$ \texttt{finish}
\par}
\end{minipage}
}
\end{tabular}
\caption{
Case study on the mismatch between task correctness and context management quality.
Cases are sampled from the traces of StateLM-8B~\cite{liu2026pensieve} on BrowseComp+~\cite{chen2025browsecomp}.
Questions are condensed to save space.
}
\label{fig:credit_assignment_case}
\end{figure}

\subsection{Pilot Study}
\label{subsec:pilot_experiment}
To better understand the impact of context management tools, we branch from context management actions in existing trajectories, sample continuation trajectories, and compute the standard deviation of their final success rates.
As shown in Figure~\ref{fig:pilot_study}, this variance differs substantially across actions, suggesting that certain context management actions have a larger impact on the final outcomes and should receive more exploration budget.
For credit assignment, Figure~\ref{fig:credit_assignment_case} shows that final correctness can be misaligned with context management quality: a correct trajectory may rely on repeated retrieval, whereas an incorrect one may still perform reasonable management.
Therefore, directly assigning the final trajectory-level reward to all intermediate snapshots may reinforce inefficient context management behaviors and penalize reasonable decisions, motivating fine-grained action-level rewards.

\section{Methodology}
\label{sec:method}

In this section, we elaborate on the methodology of \sysname{}.
We first describe the design of the extended toolset and SFT data construction, followed by the reinforcement learning recipe.

\subsection{Context Management Toolset Design}
\label{subsec:tool_design}

\sysname{} builds upon the basic toolset of StateLM~\cite{liu2026pensieve} and introduces additional tools for \textit{planning, long-term memory and soft context offloading}, as detailed in Table~\ref{tab:tools}.
To support long-term memory, we introduce \texttt{memorize}, which extracts structured information about entities, timestamps, and event episodes, and builds edges among related memory items.
The model can later call \texttt{readMemory} to retrieve a target memory together with its neighbors.
For context offloading, we introduce soft-deletion tools, including \texttt{summarizeContext}, \texttt{compressContext}, and \texttt{foldHistory}.
\texttt{foldHistory} can condense historical messages into keywords and a summary, which can be recovered by calling \texttt{searchContext} with keyword queries.
We treat context offloading, memory-writing, and memory-updating tools as context editing tools, and segment trajectories into snapshots at these operations because they modify the interaction history.
Through these extensions, \sysname{} provides a comprehensive toolset for context management, empowering models to effectively manage their working context in long-horizon tasks.
\begin{table}[htbp]
\scriptsize
\centering
\begin{tabularx}{\textwidth}{p{4cm} l X}
    \toprule
    \textbf{Category} & \textbf{Tool Name} & \textbf{Function Description} \\
    \midrule
    \multirow{3}{4cm}{Perception \& Planning}
    & \texttt{analyzeText} & Calculate context length. \\
    & \texttt{checkBudget} & Check remaining token budget. \\
    & \texttt{\textcolor{deeppurple}{plan}} & Propose a concise plan. \\
    \midrule
    \multirow{4}{4cm}{Information Retrieval} 
    & \texttt{buildIndex} & Build a searchable index. \\
    & \texttt{searchContext} & Search for relevant content. \\
    & \texttt{readChunk} & Load a specific context chunk. \\
    & \texttt{\textcolor{deeppurple}{readMultiChunks}} & Batch load multiple context chunks. \\
    \midrule
    \multirow{6}{4cm}{Memory Management} 
    & \texttt{note} & Record key information into a note. \\
    & \texttt{updateNote} & Update the content of a note. \\
    & \texttt{readNote} & Load the content of a note. \\
    & \texttt{\textcolor{deeppurple}{memorize}} & Extract key information and cross-chunk relations into event memory. \\
    & \texttt{\textcolor{deeppurple}{updateMemory}} & Update a memory item. \\
    & \texttt{\textcolor{deeppurple}{readMemory}} & Load a specific memory item. \\
    \midrule
    \multirow{4}{4cm}{Context Offloading} 
    & \texttt{deleteContext} & Replace a message with a placeholder. \\
    & \texttt{\textcolor{deeppurple}{summarizeContext}} & Replace a message with a summary. \\
    & \texttt{\textcolor{deeppurple}{compressContext}} & Compress a message via a lightweight compression model, such as llmlingua-2~\cite{pan2024llmlingua}. \\
    & \texttt{\textcolor{deeppurple}{foldHistory}} & Discard all historical messages and build a searchable index. \\
    \bottomrule
\end{tabularx}
\caption{Context management tools of \sysname{}. Tools highlighted in \textcolor{deeppurple}{purple} are newly introduced.}
\label{tab:tools}
\end{table}

\subsection{Supervised Fine-Tuning Data Synthesis}
\label{subsec:sft}
Before constructing SFT data, we design a \textbf{\textit{context management harness}} based on the developed toolset.
Through carefully orchestrated rules, the harness provides the teacher model with hints and constraints at appropriate steps.
For example, it guides the teacher to inspect retrieved content with \texttt{readChunk} after search.
When the context length exceeds a predefined threshold, it restricts the model to context offloading tools for cleaning up.
These hints and constraints are used only as generation-time scaffolding and are excluded from the final SFT trajectories.
Additional details of SFT data construction are in Appendix~\ref{apx:sft_data_synthesis}.

\subsection{RL Training with Partial Rollout and Fine-Grained Credit Assignment}
\label{subsec:rl}

\paragraph{Context-Aware Partial Rollout.} Inspired by ARPO~\cite{dong2025agentic}, we introduce partial rollout, which allocates \textbf{more exploration budget} to those critical context management decisions.
To identify these critical decisions, we first calculate the context variation:
\begin{equation}
\Delta C_t^{\mathrm{cm}} = \left| \frac{\text{len}(c_{t+1}^{\mathrm{cm}}) - \text{len}(c_{t}^{\mathrm{cm}})}{\text{len}(c_{t}^{\mathrm{cm}})} \right|,
\end{equation}
and the entropy variation:
\begin{equation}
\Delta H_t^{\mathrm{cm}} = H_{t}^{\mathrm{cm}} - H_{\mathrm{initial}},
\end{equation}
for each context management operation $a_t^{\mathrm{cm}} \in \mathcal{A}_{\mathrm{cm}}$.
$H_t$ is the average entropy of $k$ generated tokens after receiving the observation at step $t$:
\begin{equation}
\begin{aligned}
&H_t^{\mathrm{cm}} = \frac{1}{k} \sum_{i=0}^{k-1} \left( -\sum_{j=1}^{V} p_{t+i, j} \log p_{t+i, j} \right),
\end{aligned}
\end{equation}
where $V$ is the vocabulary size, $p_{t+i, j}$ denotes the probability after softmax and 
$H_{\mathrm{initial}}$ is the average entropy of the first $k$ tokens at the beginning of the trajectory.
We use the initial entropy $H_{\mathrm{initial}}$ instead of the entropy of the preceding step $H_{t-1}^{\mathrm{cm}}$ as the reference, because partial rollout aims to identify critical tool calls that induce significant uncertainty changes relative to the initial query state, instead of local fluctuations between adjacent steps.
Based on these metrics, the sensitivity score $\mathcal{S}(a_t^{\mathrm{cm}})$ can be defined as:
\begin{equation}
\mathcal{S}(a_t^{\mathrm{cm}}) = \alpha \cdot \Delta C_t^{\mathrm{cm}} + \beta \cdot \Delta H_t^{\mathrm{cm}}, 
\end{equation}
where $\alpha$ and $\beta$ balance the weights of the two metrics.

During the rollout phase, given a global rollout budget of $N$ snapshots per query, we first perform trajectory-level rollouts and segment the resulting trajectories into $M$ snapshots at context management actions.
If $M < N$, the remaining snapshot budget is allocated to partial rollouts: we rank all context management actions by their sensitivity score $\mathcal{S}(a_t^{\mathrm{cm}})$ in descending order and select the top $N - M$ actions as branching points.
Starting from the parent node of each selected action, we sample additional sub-trajectories to enrich the snapshot pool.
This adaptive partial rollout mechanism facilitates more exploration at critical context management actions that have a larger impact.

\paragraph{Fine-Grained Credit Assignment.}
Benefiting from the partial rollout mechanism, we can assign credit to intermediate trajectory snapshots in a more fine-grained manner.
Specifically, suppose a complete terminal trajectory $T$ is segmented into $M$ trajectory snapshots $\{S_1, S_2, \dots, S_M\}$ at context editing operations.
Unlike existing works~\cite{li2025sculptor,liu2026pensieve} that directly assign the final sparse reward $R(T)$ to all intermediate snapshots, we estimate the value of each snapshot using the reward of all its subsequent branches.
\textbf{For a terminal snapshot $S_M=T$}, the reward includes an outcome reward $R_{\mathrm{out}}$, a format reward $R_{\mathrm{fmt}}$, and a penalty item $R_{\mathrm{pen}}$:
\begin{equation}
R(S_M)=R_{\mathrm{out}} + R_{\mathrm{fmt}} + R_{\mathrm{pen}}.
\end{equation}
Here, $R_{\mathrm{out}}$ is computed by comparing the model's predicted answer with the ground truth, while $R_{\mathrm{fmt}}$ checks whether the final output can be successfully parsed.
The penalty term $R_{\mathrm{pen}}$ penalizes invalid tool invocations, such as calling \texttt{readMemory} before any memory has been constructed, as well as context-length violations.
\textbf{For an intermediate trajectory snapshot $S_i$}, its reward is determined by all terminal trajectories that take $S_i$ as a prefix:
\begin{equation}
R(S_i) = \frac{1}{|\mathcal{T}(S_i)|} \sum_{T \in \mathcal{T}(S_i)} R(T)
\end{equation}
where $\mathcal{T}(S_i)$ denotes the set of terminal trajectories with $S_i$ as a prefix.

After obtaining snapshot-level rewards, we group all trajectory snapshots generated under the same query $q$ as an advantage calculation group $\mathcal{G} = \{S_t^{(j)} \mid \forall j, t\}$.
We then compute the advantage of each snapshot sample $S_t^{(j)}$ using the group mean and standard deviation:
\begin{equation}
\hat{A}_t^{(j)} = \frac{R(S_t^{(j)}) - \mathrm{mean}(\{R(S) \mid S \in \mathcal{G}\})}{\mathrm{std}(\{R(S) \mid S \in \mathcal{G}\})}
\end{equation}

Finally, we optimize $\pi_\theta$ with the GRPO objective~\cite{shao2024deepseekmath}, treating each trajectory snapshot as an independent sample.
This shifts credit assignment from trajectories to snapshots, enabling more precise reward estimation for intermediate context management actions.
Appendix~\ref{apx:theory} provides a theoretical discussion of its variance reduction effect.

\section{Experiments}
\label{sec:exp}

\subsection{Experiment Setup}
\label{subsec:exp_setup}

\paragraph{Base Models and Training Data.} To evaluate the effectiveness of \sysname{}, we conduct extensive experiments on two long-horizon tasks: long-context QA and deep search.
\textbf{For long-context QA}, we follow the data construction setup of StateLM~\cite{liu2026pensieve}.
Specifically, we construct SFT data from the PublicDomain split of NovelQA~\cite{wangnovelqa} and the training split of NarrativeQA~\cite{kovcisky2018narrativeqa}, and perform reinforcement learning on the training set of LongBench-v2~\cite{bai2025longbench}.
For model selection, we use Qwen3-8B, Qwen3-14B~\cite{yang2025qwen3} and Gemma4-E4B-it~\cite{gemma4} as base models.
\textbf{For deep search}, base models are WebSailor-7B~\cite{li2025websailor} and WebExplorer-8B~\cite{liu2025webexplorer}.
Since they already possess basic search capabilities, we skip SFT and directly perform RL training on $1$K samples drawn from OpenSeeker~\cite{du2026openseeker}.
Detailed statistics of training data are provided in Appendix Table~\ref{tab:training_data_stats}.

\paragraph{Implementation Details.} We conduct SFT and RL training with the verl library~\cite{sheng2025hybridflow}.
In RL, we limit each complete trajectory to be segmented into at most $8$ trajectory snapshots.
For each query, we collect $N=128$ trajectory snapshots, starting from $8$ trajectory-level rollouts, producing at most $M=64$ trajectory snapshots, and completing the remaining budget with partial rollout.
During inference, we set the maximum input and generation lengths to $30$K and $2$K tokens, respectively.
More details are in Appendix~\ref{apx:train_details}.

\paragraph{Evaluation Benchmarks and Metrics.} We evaluate long-context QA on four 
benchmarks: the Copyright split of NovelQA~\cite{wangnovelqa}, the En.MC split of $\infty$Bench~\cite{zhang2024infty}, LongMemEval-S~\cite{wu2025longmemeval}, and BrowseComp+~\cite{chen2025browsecomp}. 
Note that BrowseComp+ is built on a fixed corpus and does not require searching on the Internet.
We therefore include it in the long-context QA tasks.
The evaluation on deep search tasks includes GAIA (the text-only subset with $103$ examples)~\cite{mialon2024gaia}, BrowseComp~\cite{bc_en}, BrowseComp-ZH~\cite{bc_zh}, and xBench-DeepSearch~\cite{xbench}.
We adopt exact-match evaluation for multiple-choice benchmarks, NovelQA and $\infty$Bench.
Other benchmarks are evaluated by LLM-as-a-Judge.

\paragraph{Baselines.}
\textbf{For long-context QA}, we compare \sysname{} with 
(1) a training-free method: ReadAgent~\cite{lee2024human}, 
(2) an RL-trained memory agent: MemAgent~\cite{yu2025memagent}, 
(3) a proactive context management agent: StateLM~\cite{liu2026pensieve},
and (4) a prompt-only baseline that uses the same tools as \sysname{} but without any fine-tuning, denoted as ``w/ tools''.
\textbf{Regarding deep search}, we include the following baselines: 
(1) ReAct (w/ truncation), which truncates early messages when the context exceeds $28$K tokens,
(2) ReSum~\cite{wu2025resum}, an inference-time summarization method, 
(3) SUPO~\cite{lu2025scaling}, which jointly trains summarization and agentic ability through RL, 
and (4) OpenSeeker, which performs RL training on the same $1$K samples as ours but without context management tools.
Baseline replication details are in Appendix~\ref{apx:baseline_details}.

\begin{table*}[t!]
\centering
\caption{Performance comparison of \sysname{} against baseline methods on long-context QA tasks.
We run each method three times and report the mean and standard deviation.
Results with \textsuperscript{$\dagger$} are from~\protect\citet{liu2026pensieve}.}
\label{tab:longdoc_qa_results}
\small
\def\arraystretch{1.12}
\resizebox{\textwidth}{!}{
\begin{tabular}{lcccccc}
\toprule
\rowcolor{blue!15}
\textbf{Model} & \textbf{Length} & \textbf{NovelQA} & \textbf{$\infty$Bench} & \textbf{LongMemEval-S} & \textbf{BrowseComp+} & \textbf{Avg.} \\
\midrule
Qwen3.5-397B-A17B (w/o tools) & 256K & 88.77 & 90.39 & 81.00 & 62.05 & 80.55 \\
Qwen3.5-397B-A17B (w/ tools) & 32K & 91.94 & 92.13 & 83.60 & 80.96 & 87.16 \\
\midrule
RL-MemoryAgent-7B\textsuperscript{$\dagger$} & 32K & 60.24 & 62.45 & 40.60 & - & - \\
RL-MemoryAgent-14B\textsuperscript{$\dagger$} & 32K & 78.86 & 74.24 & 59.00 & - & - \\
ReadAgent-8B\textsuperscript{$\dagger$} & 32K & 16.38 & 24.02 & 0.00 & - & - \\
ReadAgent-14B\textsuperscript{$\dagger$} & 32K & 23.12 & 34.06 & 14.60 & - & - \\
StateLM-8B-RL\textsuperscript{$\dagger$} & 32K & 84.15 {\textnormal{\scriptsize $\pm$ 1.00}} & 73.07 {\textnormal{\scriptsize $\pm$ 1.33}} & 59.73 {\textnormal{\scriptsize $\pm$ 2.20}} & 46.44 {\textnormal{\scriptsize $\pm$ 0.77}} & 65.85 \\
StateLM-14B-RL\textsuperscript{$\dagger$} & 32K & 84.85 {\textnormal{\scriptsize $\pm$ 0.42}} & 78.46 {\textnormal{\scriptsize $\pm$ 0.67}} & 64.47 {\textnormal{\scriptsize $\pm$ 0.50}} & 52.67 {\textnormal{\scriptsize $\pm$ 4.00}} & 70.11 \\
\midrule
Qwen3-8B (w/o tools) & 128K & 65.74 {\textnormal{\scriptsize $\pm$ 0.55}} & 66.96 {\textnormal{\scriptsize $\pm$ 1.09}} & 45.20 {\textnormal{\scriptsize $\pm$ 1.02}} & 5.82 {\textnormal{\scriptsize $\pm$ 0.86}} & 45.93 \\
Qwen3-8B (w/ tools) & 32K & 38.09 {\textnormal{\scriptsize $\pm$ 0.88}} & 39.59 {\textnormal{\scriptsize $\pm$ 1.03}} & 24.47 {\textnormal{\scriptsize $\pm$ 0.57}} & 8.28 {\textnormal{\scriptsize $\pm$ 0.11}} & 27.61 \\
\rowcolor{cyan!6}
\textbf{ContextPilot-8B} & 32K & 82.56 {\textnormal{\scriptsize $\pm$ 0.49}} & 71.03 {\textnormal{\scriptsize $\pm$ 1.44}} & 60.67 {\textnormal{\scriptsize $\pm$ 1.91}} & 48.84 {\textnormal{\scriptsize $\pm$ 1.48}} & 65.78 \\
\rowcolor{cyan!6}
\textbf{ContextPilot-8B-RL} & 32K & \textbf{83.88} {\textnormal{\scriptsize $\pm$ 0.67}} & \textbf{75.25} {\textnormal{\scriptsize $\pm$ 0.82}} & \textbf{64.27} {\textnormal{\scriptsize $\pm$ 1.15}} & \textbf{54.18} {\textnormal{\scriptsize $\pm$ 1.47}} & \textbf{69.40} \\
\midrule
Qwen3-14B (w/o tools) & 128K & 78.03 {\textnormal{\scriptsize $\pm$ 0.44}} & 74.53 {\textnormal{\scriptsize $\pm$ 0.41}} & 54.20 {\textnormal{\scriptsize $\pm$ 0.75}} & 6.27 {\textnormal{\scriptsize $\pm$ 0.87}} & 53.26 \\
Qwen3-14B (w/ tools) & 32K & 68.43 {\textnormal{\scriptsize $\pm$ 1.22}} & 54.59 {\textnormal{\scriptsize $\pm$ 0.94}} & 40.33 {\textnormal{\scriptsize $\pm$ 0.50}} & 15.78 {\textnormal{\scriptsize $\pm$ 0.10}} & 44.78 \\
\rowcolor{cyan!6}
\textbf{ContextPilot-14B} & 32K & 84.28 {\textnormal{\scriptsize $\pm$ 0.76}} & 79.04 {\textnormal{\scriptsize $\pm$ 0.94}} & 65.93 {\textnormal{\scriptsize $\pm$ 1.20}} & 53.13 {\textnormal{\scriptsize $\pm$ 1.37}} & 70.60 \\
\rowcolor{cyan!6}
\textbf{ContextPilot-14B-RL} & 32K & \textbf{84.81} {\textnormal{\scriptsize $\pm$ 0.86}} & \textbf{81.08} {\textnormal{\scriptsize $\pm$ 1.15}} & \textbf{67.40} {\textnormal{\scriptsize $\pm$ 0.91}} & \textbf{55.50} {\textnormal{\scriptsize $\pm$ 1.71}} & \textbf{72.20} \\
\midrule
Gemma4-E4B-it (w/o tools) & 128K & 48.75 {\textnormal{\scriptsize $\pm$ 0.53}} & 39.74 {\textnormal{\scriptsize $\pm$ 0.87}} & 28.50 {\textnormal{\scriptsize $\pm$ 1.90}} & 7.03 {\textnormal{\scriptsize $\pm$ 0.30}} & 31.01 \\
Gemma4-E4B-it (w/ tools) & 32K & 36.90 {\textnormal{\scriptsize $\pm$ 2.06}} & 32.75 {\textnormal{\scriptsize $\pm$ 2.17}} & 21.80 {\textnormal{\scriptsize $\pm$ 1.30}} & 2.73 {\textnormal{\scriptsize $\pm$ 0.30}} & 23.55 \\
\rowcolor{cyan!6}
\textbf{ContextPilot-E4B} & 32K & 66.80 {\textnormal{\scriptsize $\pm$ 0.81}} & 55.02 {\textnormal{\scriptsize $\pm$ 1.07}} & 55.07 {\textnormal{\scriptsize $\pm$ 1.20}} & 42.05 {\textnormal{\scriptsize $\pm$ 1.04}} & 54.74 \\
\rowcolor{cyan!6}
\textbf{ContextPilot-E4B-RL} & 32K & \textbf{72.92} {\textnormal{\scriptsize $\pm$ 0.86}} & \textbf{60.99} {\textnormal{\scriptsize $\pm$ 1.25}} & \textbf{62.47} {\textnormal{\scriptsize $\pm$ 1.62}} & \textbf{47.47} {\textnormal{\scriptsize $\pm$ 1.08}} & \textbf{60.96} \\
\bottomrule
\end{tabular}
}
\end{table*}

\begin{table*}[t!]
\centering
\caption{Performance comparison of \sysname{} against baseline methods on deep search tasks.
We run each method three times and report the mean and standard deviation.}
\label{tab:deep_research_results}
\small
\def\arraystretch{1.12}
\resizebox{\textwidth}{!}{
\begin{tabular}{llccccc}
\toprule
\rowcolor{blue!15}
\textbf{Backbone} & \textbf{Method} & \textbf{BrowseComp} & \textbf{BrowseComp-ZH} & \textbf{GAIA} & \textbf{xBench-DS} & \textbf{Avg.} \\
\rowcolor{blue!15}
& & \textbf{pass@3} & \textbf{pass@3} & \textbf{pass@1} & \textbf{pass@1} & \\
\midrule
\multirow{6}{*}{WebSailor-7B} & ReAct & 11.33 {\textnormal{\scriptsize $\pm$ 1.03}} & 25.47 {\textnormal{\scriptsize $\pm$ 0.98}} & 31.07 {\textnormal{\scriptsize $\pm$ 0.79}} & 34.00 {\textnormal{\scriptsize $\pm$ 0.82}} & 25.47 \\
& ReAct (w/ truncation) & 12.67 {\textnormal{\scriptsize $\pm$ 0.62}} & 27.91 {\textnormal{\scriptsize $\pm$ 0.91}} & 33.66 {\textnormal{\scriptsize $\pm$ 0.92}} & 35.00 {\textnormal{\scriptsize $\pm$ 0.82}} & 27.31 \\
& ReSum & 15.83 {\textnormal{\scriptsize $\pm$ 0.85}} & 38.99 {\textnormal{\scriptsize $\pm$ 1.56}} & 38.19 {\textnormal{\scriptsize $\pm$ 1.21}} & 35.33 {\textnormal{\scriptsize $\pm$ 1.25}} & 32.09 \\
& SUPO & 18.50 {\textnormal{\scriptsize $\pm$ 1.08}} & 42.68 {\textnormal{\scriptsize $\pm$ 0.99}} & 42.07 {\textnormal{\scriptsize $\pm$ 1.65}} & 42.00 {\textnormal{\scriptsize $\pm$ 1.41}} & 36.31 \\
& OpenSeeker & 16.50 {\textnormal{\scriptsize $\pm$ 2.12}} & 41.87 {\textnormal{\scriptsize $\pm$ 1.02}} & 41.42 {\textnormal{\scriptsize $\pm$ 1.21}} & 43.33 {\textnormal{\scriptsize $\pm$ 0.94}} & 35.78 \\
\rowcolor{cyan!6}
& \textbf{ContextPilot} & \textbf{21.17} {\textnormal{\scriptsize $\pm$ 1.31}} & \textbf{43.14} {\textnormal{\scriptsize $\pm$ 1.07}} & \textbf{45.31} {\textnormal{\scriptsize $\pm$ 0.92}} & \textbf{43.67} {\textnormal{\scriptsize $\pm$ 0.94}} & \textbf{38.32} \\
\midrule
\multirow{6}{*}{WebExplorer-8B} & ReAct & 23.83 {\textnormal{\scriptsize $\pm$ 1.55}} & 46.57 {\textnormal{\scriptsize $\pm$ 0.74}} & 48.87 {\textnormal{\scriptsize $\pm$ 0.92}} & 52.33 {\textnormal{\scriptsize $\pm$ 0.94}} & 42.90 \\
& ReAct (w/ truncation) & 24.33 {\textnormal{\scriptsize $\pm$ 1.18}} & 45.91 {\textnormal{\scriptsize $\pm$ 1.07}} & 49.84 {\textnormal{\scriptsize $\pm$ 0.46}} & 52.33 {\textnormal{\scriptsize $\pm$ 0.47}} & 43.10 \\
& ReSum & 28.83 {\textnormal{\scriptsize $\pm$ 1.03}} & 47.87 {\textnormal{\scriptsize $\pm$ 1.39}} & 52.75 {\textnormal{\scriptsize $\pm$ 1.21}} & 51.00 {\textnormal{\scriptsize $\pm$ 1.41}} & 45.11 \\
& SUPO & 31.00 {\textnormal{\scriptsize $\pm$ 1.78}} & 50.40 {\textnormal{\scriptsize $\pm$ 1.07}} & 56.96 {\textnormal{\scriptsize $\pm$ 1.21}} & \textbf{58.00} {\textnormal{\scriptsize $\pm$ 0.82}} & 49.09 \\
& OpenSeeker & 29.17 {\textnormal{\scriptsize $\pm$ 1.31}} & 48.56 {\textnormal{\scriptsize $\pm$ 1.39}} & 57.28 {\textnormal{\scriptsize $\pm$ 1.37}} & 56.67 {\textnormal{\scriptsize $\pm$ 0.47}} & 47.92 \\
\rowcolor{cyan!6}
& \textbf{ContextPilot} & \textbf{32.17} {\textnormal{\scriptsize $\pm$ 1.18}} & \textbf{53.63} {\textnormal{\scriptsize $\pm$ 1.23}} & \textbf{57.93} {\textnormal{\scriptsize $\pm$ 0.92}} & 56.67 {\textnormal{\scriptsize $\pm$ 0.94}} & \textbf{50.10} \\
\bottomrule
\end{tabular}
}
\end{table*}

\subsection{Main Results}

Tables~\ref{tab:longdoc_qa_results} and~\ref{tab:deep_research_results} compare \sysname{} with various baselines on long-context QA and deep search tasks.
We highlight three key observations.

\noindent\textbf{(1) \sysname{} achieves the best average performance among comparable-size models.}
Shown by Table~\ref{tab:longdoc_qa_results}, \sysname{} uses only a $32$K context window, yet outperforms the $128$K backbone.
\sysname{} also surpasses prior RL-trained context management agents.
For example, ContextPilot-8B-RL outperforms StateLM-8B-RL by an average of $3.55$ points across four benchmarks; on deep search tasks, ContextPilot also surpasses SUPO by $1.51$ average points across both backbones.

\noindent\textbf{(2) RL training brings more pronounced gains in more complex long-context settings.}
Starting from the SFT model, RL further improves \sysname{}.
For instance, ContextPilot-8B-RL improves over ContextPilot-8B by an average of $3.62$ points across the four long-context QA tasks.
The improvement on NovelQA is relatively modest, because the SFT data already includes another split of NovelQA.
By contrast, RL yields a substantial $5.34$ point increase on the more challenging BrowseComp+ benchmark.
Given that the average input length of NovelQA is about $119$K tokens whereas BrowseComp+ reaches $552$K tokens, this contrast suggests that the benefit of RL is larger on more challenging tasks.

\noindent\textbf{(3) The advantages of \sysname{} generalize across tasks and base models.}
As shown in Tables~\ref{tab:longdoc_qa_results} and~\ref{tab:deep_research_results}, \sysname{} brings consistent gains on both long-context QA and deep search.
The gains also hold across different backbones, including Qwen3-8B, Qwen3-14B and Gemma4-E4B on long-context QA, as well as WebSailor-7B and WebExplorer-8B on deep search.

\subsection{Further Analysis}

\begin{wrapfigure}{r}{0.48\columnwidth}
    \centering
    \vspace{-5mm}
    \includegraphics[width=\linewidth]{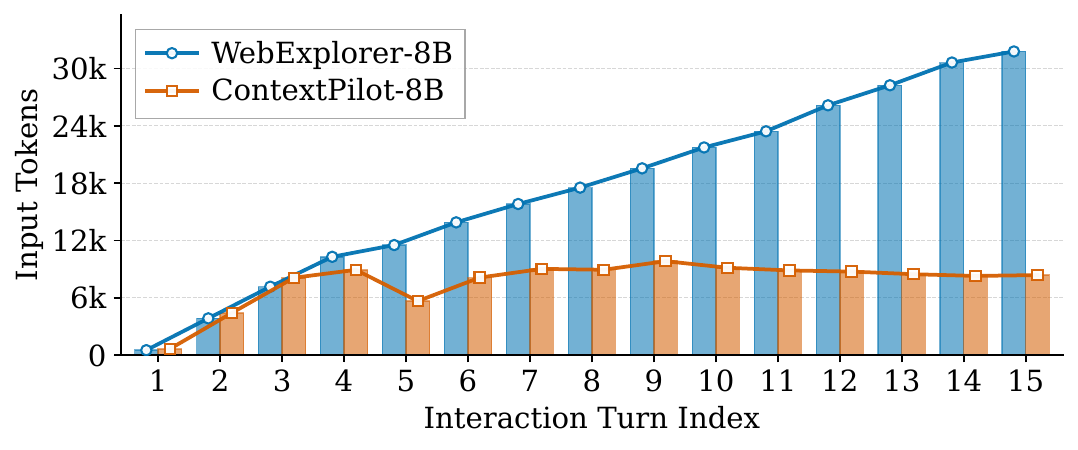}
    \vspace{-7mm}
    \caption{Token usage per turn on BrowseComp.}
    \label{fig:token_count}
    \vspace{-10mm}
\end{wrapfigure}

\paragraph{Token efficiency analysis.}
Beyond performance, we examine whether \sysname{} can maintain a more compact working context.
We consider trajectories with at least $15$ turns and compute the average number of input tokens per turn.
As shown in Figure~\ref{fig:token_count}, the input length of WebExplorer-8B grows almost linearly on BrowseComp, reaching around $30$K tokens.
By contrast, ContextPilot-8B stabilizes its input length each turn at roughly $8$K--$10$K tokens.

\begin{figure*}[t]
    \centering
    \begin{subfigure}{0.49\linewidth}
        \centering
        \includegraphics[width=\linewidth]{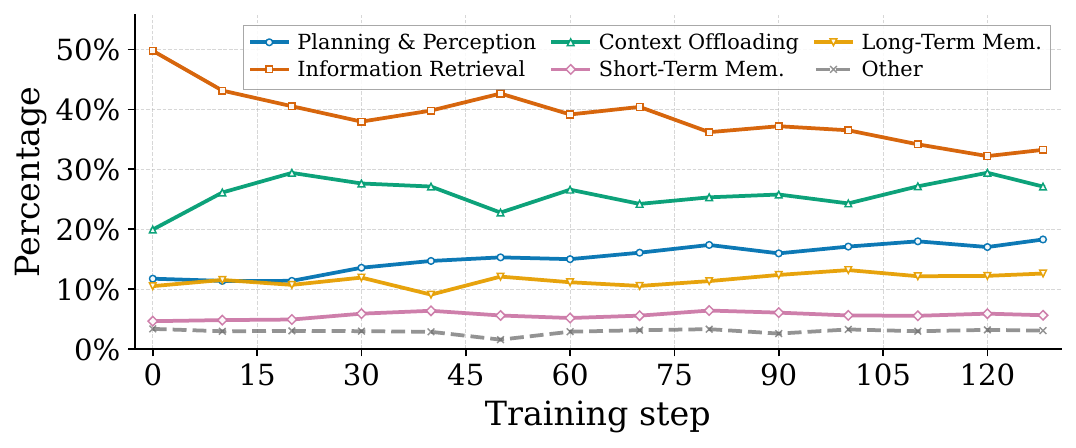}
        \caption{$\infty$Bench.}
        \label{fig:tool_statics_infbench}
    \end{subfigure}
    \hfill
    \begin{subfigure}{0.49\linewidth}
        \centering
        \includegraphics[width=\linewidth]{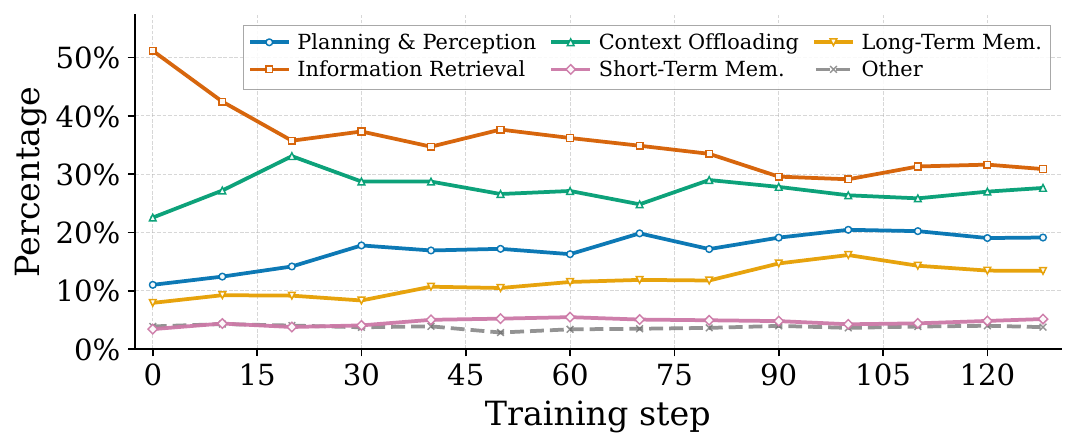}
        \caption{NovelQA.}
        \label{fig:tool_statics_novelqa}
    \end{subfigure}
    \caption{Tool category breakdown during Qwen3-8B RL training on $\infty$Bench and NovelQA.}
    \label{fig:tool_statics}
\end{figure*}

\paragraph{RL reshapes tool use strategy.}
To better understand how RL changes the context management behavior, we track the distribution over tool call categories of ContextPilot-8B-RL in RL training.
As shown in Figure~\ref{fig:tool_statics}, the model relies heavily on information retrieval tools in the early stage of RL training, with retrieval accounting for roughly half of all tool calls.
As RL training proceeds, the share of information retrieval tools gradually decreases, whereas planning and perception, long-term memory, and context offloading tools exhibit an upward trend.
This shift suggests that RL encourages the model to move beyond naive information retrieval toward more coordinated context management.

\begin{figure*}[t]
    \centering
    \begin{subfigure}{0.49\linewidth}
        \centering
        \includegraphics[width=\linewidth]{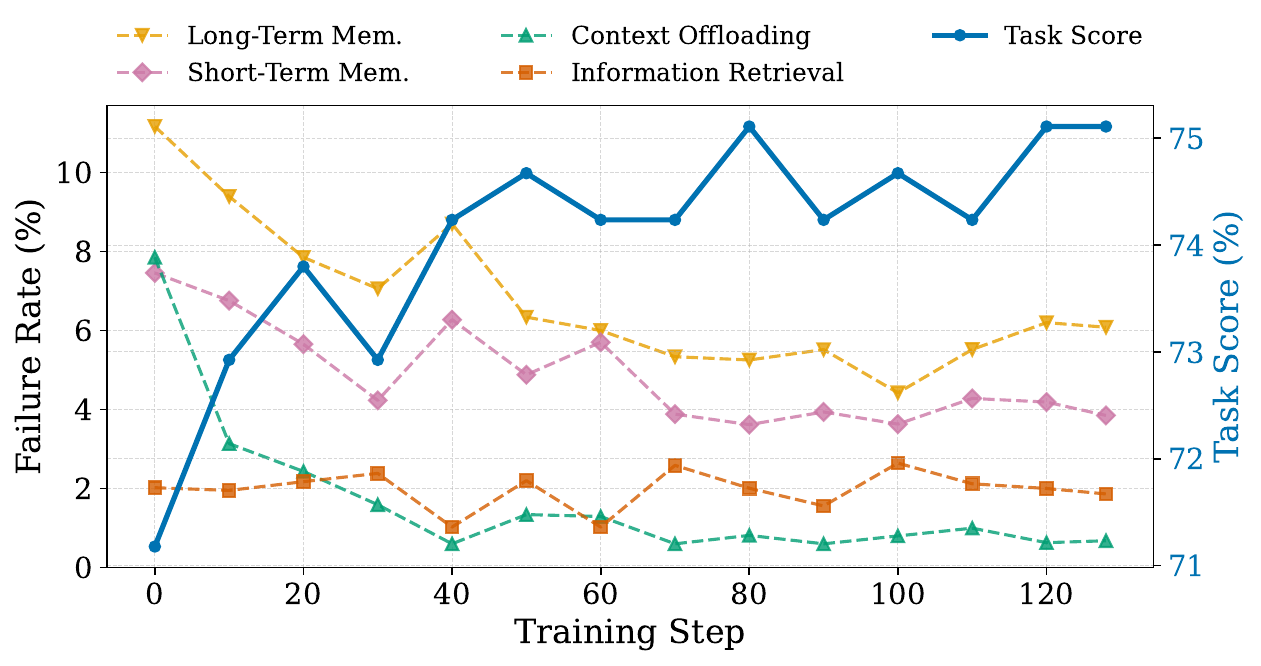}
        \caption{$\infty$Bench.}
        \label{fig:tool_error_infbench}
    \end{subfigure}
    \hfill
    \begin{subfigure}{0.49\linewidth}
        \centering
        \includegraphics[width=\linewidth]{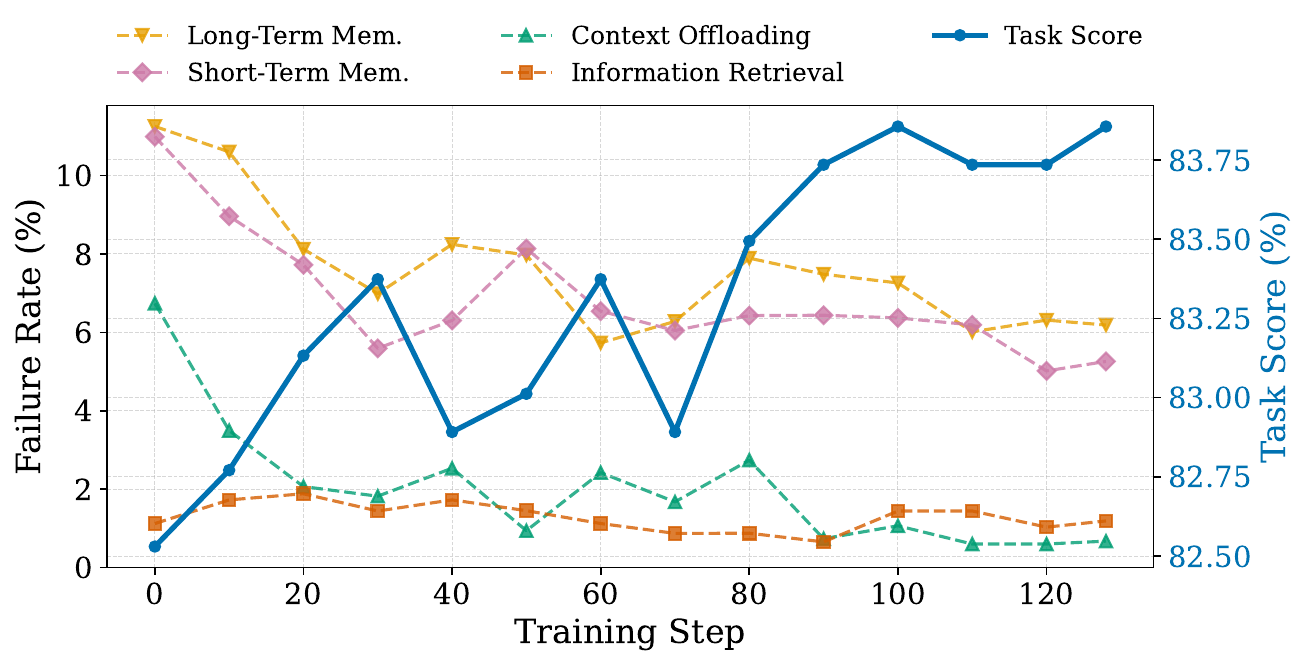}
        \caption{NovelQA.}
        \label{fig:tool_error_novelqa}
    \end{subfigure}
    \caption{Tool invocation failure rates and task success rate on $\infty$Bench and NovelQA during RL training.}
    \label{fig:tool_error}
\end{figure*}

\paragraph{RL synergistically boosts tool-use correctness and task success.}
Beyond the tool call frequency breakdown, we further examine whether RL improves the correctness of context management tool use.
To quantify this effect, we define a tool invocation as failed when it triggers an environment-side error---such as a malformed call, invalid arguments, or a violation of tool-specific preconditions---and count such failures for each intermediate checkpoint of ContextPilot-8B-RL.
Figure~\ref{fig:tool_error} reports the failure rates of four representative tool categories together with the task success rate throughout RL training.
At the early stage of training, memory and context offloading tools exhibit substantially higher failure rates than information retrieval tools.
This pattern suggests that although the SFT model can invoke these tools, it lacks a genuine understanding of their usage.
During the RL process, the model's proficiency with context management tools contributes to the improvement in task success rate.

\subsection{Ablation Studies}

\begin{wraptable}{r}{0.5\columnwidth}
\centering
\vspace{-4mm}
\caption{Ablation results of tool design using Qwen3.5-397B-A17B. LME-S and BC+ denote LongMemEval-S and BrowseComp+, respectively.}
\label{tab:tool_design_ablation}
\small
\setlength{\tabcolsep}{3pt}
\def\arraystretch{1.12}
\resizebox{0.5\columnwidth}{!}{
\begin{tabular}{lccccc}
\toprule
\textbf{Tool Design} & \textbf{NovelQA} & \textbf{$\infty$Bench} & \textbf{LME-S} & \textbf{BC+} & \textbf{Avg.} \\
\midrule
Original tools & 88.90 & 85.15 & 74.00 & 63.49 & 77.89 \\
+ Planning & 89.76 & 87.23 & 78.50 & 65.66 & 80.29 \\
+ Soft offloading & 91.28 & 89.63 & 80.20 & 71.20 & 83.08 \\
+ Long-term memory & \textbf{91.94} & \textbf{92.13} & \textbf{83.60} & \textbf{80.96} & \textbf{87.16} \\
\bottomrule
\end{tabular}
}
\vspace{-4mm}
\end{wraptable}

\paragraph{Tool design.}
To isolate the effect of tool design, we evaluate Qwen3.5-397B-A17B with cumulative tool configurations, progressively adding planning, soft context offloading, and long-term memory tools.
As presented in Table~\ref{tab:tool_design_ablation}, the performance improves as more context management tools are introduced, with the full toolset achieving the highest score.
The gains are especially clear on BrowseComp+, where accuracy increases from $63.49\%$ to $80.96\%$ with the full toolset, highlighting the effectiveness of the newly introduced tools.

\begin{wraptable}{r}{0.5\columnwidth}
\centering
\caption{Ablation results of the RL algorithm using Qwen3-8B. 
Results for Gemma4-E4B-it are in Table~\ref{tab:rl_training_ablation_gemma} (Appendix).
``+ Entropy.'', ``+ Context.'', and ``+ Fine-grained.'' 
denote RL training with entropy-based partial rollout, context-aware partial rollout, and fine-grained credit assignment.
Numbers in parentheses indicate changes relative to the previous line.
}
\label{tab:rl_training_ablation}
\scriptsize
\setlength{\tabcolsep}{2pt}
\def\arraystretch{1.12}
\resizebox{0.5\columnwidth}{!}{
\begin{tabular}{lcccc}
\toprule
\textbf{Training method} & \textbf{NovelQA} & \textbf{$\infty$Bench} & \textbf{LME-S} & \textbf{BC+} \\
\midrule
SFT & 82.56 & 71.03 & 60.67 & 48.84 \\
GRPO & 83.53 {\scriptsize \textcolor{green2}{(+0.97)}} & 72.78 {\scriptsize \textcolor{green2}{(+1.75)}} & 60.07 {\scriptsize \textcolor{red}{(-0.60)}} & 50.96 {\scriptsize \textcolor{green2}{(+2.12)}} \\
+ Entropy. & 82.52 {\scriptsize \textcolor{red}{(-1.01)}} & 73.07 {\scriptsize \textcolor{green2}{(+0.29)}} & 62.13 {\scriptsize \textcolor{green2}{(+2.06)}} & 49.64 {\scriptsize \textcolor{red}{(-1.32)}} \\
+ Context. & 83.05 {\scriptsize \textcolor{green2}{(+0.53)}} & 73.94 {\scriptsize \textcolor{green2}{(+0.87)}} & 61.40 {\scriptsize \textcolor{red}{(-0.73)}} & 51.08 {\scriptsize \textcolor{green2}{(+1.44)}} \\
+ Fine-grained. & \textbf{83.88} {\scriptsize \textcolor{green2}{(+0.83)}} & \textbf{75.25} {\scriptsize \textcolor{green2}{(+1.31)}} & \textbf{64.27} {\scriptsize \textcolor{green2}{(+2.87)}} & \textbf{54.18} {\scriptsize \textcolor{green2}{(+3.10)}} \\
\bottomrule
\end{tabular}
}
\vspace{-4mm}
\end{wraptable}

\paragraph{RL training design.}
We further ablate context-aware partial rollout and fine-grained credit assignment in RL training.
As shown in Table~\ref{tab:rl_training_ablation}, entropy-based partial rollout improves several tasks but remains unstable, decreasing BrowseComp+ accuracy by $1.32$ points on Qwen3-8B.
This suggests that entropy alone is insufficient for identifying critical context editing operations, while adding context variation yields more stable gains.
Fine-grained credit assignment further improves over context-aware partial rollout across all four benchmarks, highlighting the importance of assigning more granular action-level credit rather than relying only on trajectory-level rewards.

\section{Conclusion}

We present \sysname{}, a proactive context management agent system for agentic reasoning.
It extends the context management toolset with planning, long-term memory, and soft context offloading tools, and further improves RL training with context-aware partial rollout and fine-grained credit assignment.
Experiments show that \sysname{} maintains a more compact working context while achieving stronger performance, bringing consistent gains on both long-context QA and deep search tasks.
These results highlight the importance of proactive context management and point to a promising direction for building stronger agent systems that can self-manage their context throughout long-horizon tasks.

\section*{Limitations}

While \sysname{} demonstrates strong performance and improved token efficiency, this work has several limitations.
First, although our approach extends existing context management tools, the toolset may still not cover all forms of context editing demands.
Future work can explore richer operations for organizing, compressing, and retrieving context under different task requirements.
Second, due to computational constraints, we do not conduct extensive search over some training hyperparameters.
These settings may affect training efficiency and final performance, especially for partial rollout and credit assignment.
Finally, our experiments focus mainly on long-context QA and deep search tasks.
Extending proactive context management to broader agentic reasoning scenarios, such as agentic coding and GUI agents, remains an important direction for future work.

\setcitestyle{numbers,square}
\setcitestyle{square,numbers,comma}
\bibliography{youtu_bib}

\appendix
\section{Theoretical Discussion of Fine-Grained Credit Assignment}
\label{apx:theory}

We justify our fine-grained credit assignment as a lower-variance estimator of the conditional continuation value.
For a query $q$, let $T$ be a complete trajectory with terminal reward $R(T)$, and let $S$ be an intermediate trajectory snapshot induced by a context editing action.
The target credit for $S$ is
\begin{equation}
\label{eq:target_credit}
    Q(S)
    \triangleq
    \mathbb{E}\!\left[
        R(T)
        \mid S \preceq T
    \right],
\end{equation}
where $S \preceq T$ denotes that $S$ is a prefix snapshot of $T$.

Trajectory-level credit assignment uses the reward of the single terminal trajectory containing $S$, while our method averages all sampled terminal continuations that pass through $S$:
\begin{equation}
\label{eq:credit_estimators}
\begin{aligned}
    \widehat{Q}_{\mathrm{traj}}(S)
    &\triangleq
    R(T_1), \\
    \widehat{Q}_{\mathrm{ours}}(S)
    &\triangleq
    \frac{1}{n_S}
    \sum_{k=1}^{n_S}
    R(T_k).
\end{aligned}
\end{equation}
Here $n_S=|\mathcal{T}(S)|$, and each $T_k$ is sampled from the same continuation distribution $p(\cdot\mid S)$.

\paragraph{Claim.}
Conditioned on $S$, suppose $\{R(T_k)\}_{k=1}^{n_S}$ are independent samples with finite conditional variance
\begin{equation}
\label{eq:conditional_variance}
    \sigma^2(S)
    \triangleq
    \operatorname{Var}\!\left[
        R(T)
        \mid S \preceq T
    \right].
\end{equation}
Then both estimators are unbiased:
\begin{equation}
\label{eq:unbiasedness}
\begin{aligned}
    \mathbb{E}\!\left[
        \widehat{Q}_{\mathrm{traj}}(S)
        \mid S
    \right]
    &=
    Q(S), \\
    \mathbb{E}\!\left[
        \widehat{Q}_{\mathrm{ours}}(S)
        \mid S
    \right]
    &=
    Q(S),
\end{aligned}
\end{equation}
and their conditional variances satisfy
\begin{equation}
\label{eq:variance_reduction}
\begin{aligned}
    \operatorname{Var}\!\left[
        \widehat{Q}_{\mathrm{traj}}(S)
        \mid S
    \right]
    &=
    \sigma^2(S), \\
    \operatorname{Var}\!\left[
        \widehat{Q}_{\mathrm{ours}}(S)
        \mid S
    \right]
    &=
    \frac{\sigma^2(S)}{n_S}.
\end{aligned}
\end{equation}
Therefore, when $n_S>1$ and $\sigma^2(S)>0$, our estimator has strictly smaller mean-squared error than trajectory-level credit assignment.

\begin{proof}
Since each continuation is sampled from $p(\cdot\mid S)$,
\begin{equation}
\begin{aligned}
    \mathbb{E}\!\left[
        \widehat{Q}_{\mathrm{traj}}(S)
        \mid S
    \right]
    &=
    \mathbb{E}\!\left[
        R(T_1)
        \mid S
    \right] \\
    &=
    Q(S),
\end{aligned}
\end{equation}
and
\begin{equation}
\begin{aligned}
    \mathbb{E}\!\left[
        \widehat{Q}_{\mathrm{ours}}(S)
        \mid S
    \right]
    &=
    \frac{1}{n_S}
    \sum_{k=1}^{n_S}
    \mathbb{E}\!\left[
        R(T_k)
        \mid S
    \right] \\
    &=
    Q(S).
\end{aligned}
\end{equation}
Thus both estimators are unbiased.

The trajectory-level estimator uses one continuation, so
\begin{equation}
    \operatorname{Var}\!\left[
        \widehat{Q}_{\mathrm{traj}}(S)
        \mid S
    \right]
    =
    \sigma^2(S).
\end{equation}
For our estimator, conditional independence gives
\begin{equation}
\begin{aligned}
    &\operatorname{Var}\!\left[
        \widehat{Q}_{\mathrm{ours}}(S)
        \mid S
    \right] \\
    &\quad =
    \operatorname{Var}\!\left[
        \frac{1}{n_S}
        \sum_{k=1}^{n_S}
        R(T_k)
        \,\middle|\, S
    \right] \\
    &\quad =
    \frac{1}{n_S^2}
    \sum_{k=1}^{n_S}
    \operatorname{Var}\!\left[
        R(T_k)
        \mid S
    \right] \\
    &\quad =
    \frac{\sigma^2(S)}{n_S}.
\end{aligned}
\end{equation}
Since both estimators are unbiased, their mean-squared errors equal their variances.
The claim follows.
\end{proof}

Thus, our method estimates the same target credit $Q(S)$ as trajectory-level credit assignment with lower variance.
Since GRPO computes relative advantages from these reward estimates within each query group, this lower-variance estimator provides a more stable advantage signal for optimizing context editing decisions.
Context-aware partial rollout strengthens this effect by increasing $n_S$ for sensitive context management actions.

\section{Training Dataset Statistics}
\label{apx:training_data_stats}

Table~\ref{tab:training_data_stats} summarizes the training data used by \sysname{} across SFT and RL stages, for both experimental scenarios.
For SFT, we construct trajectories from the PublicDomain split of NovelQA and the training split of NarrativeQA, and retain trajectories that satisfy the outcome and context management quality requirements.
The retained trajectories are further segmented into trajectory snapshots at context editing operations, producing $51{,}469$ snapshots in total.
For RL, we use $488$ LongBench-v2 questions for long-context QA and $1{,}000$ questions randomly sampled from the OpenSeeker~\cite{du2026openseeker} dataset for deep search.

\section{SFT Data Synthesis Details}
\label{apx:sft_data_synthesis}

We synthesize SFT trajectories for long-context QA only.
The teacher is Qwen3.5-397B-A17B~\cite{qwen35blog} in thinking mode, decoded with temperature $0.6$, top-$p$ of $0.95$, and a maximum output length of $4$K tokens.
To elicit valid context management behavior from the teacher, we use a context management harness that supplies tool definitions together with procedural guidance for managing long interaction histories.
Rather than exposing the full toolset at every step, the harness dynamically presents only tools whose preconditions are satisfied by the current state.
For instance, \texttt{readMemory} is unavailable before any memory has been written, while \texttt{readChunk} and \texttt{readMultiChunks} are withheld until \texttt{searchContext} has been invoked.
The harness also implements correction and retry mechanisms for recoverable generation errors.
When the teacher provides an invalid argument, such as a nonexistent message id or an already deleted message id for \texttt{deleteContext}, \texttt{summarizeContext} or \texttt{compressContext}, the harness returns a targeted hint and asks the teacher to retry the tool call.
Similarly, if the teacher attempts to answer in plain text without invoking \texttt{finish}, the harness reminds it to submit the answer through the required tool interface.
These auxiliary hints serve only as generation-time scaffolding and are excluded from the final trajectories.
If a retry is triggered, we keep only the final successful attempt, preventing failed intermediate calls from being imitated.

\begin{wraptable}{r}{0.5\columnwidth}
\centering
\vspace{-4mm}
\small
\setlength{\tabcolsep}{5pt}
\resizebox{0.5\columnwidth}{!}{
\begin{tabular}{llrrr}
\toprule
\textbf{Stage} & \textbf{Source} & \textbf{Questions} & \textbf{Trajectories} & \textbf{Snapshots} \\
\midrule
\multirow{2}{*}{SFT}
& NovelQA & 3,096 & 2,987 & 50,691 \\
& NarrativeQA & 100 & 81 & 778 \\
\midrule
\multirow{2}{*}{RL}
& LongBench-v2 & 488 & -- & -- \\
& OpenSeeker-v1 (sampled) & 1,000 & -- & --  \\
\bottomrule
\end{tabular}
}
\caption{Statistics of the training data, where NovelQA, NarrativeQA, and LongBench-v2 are used for long-context QA and OpenSeeker-v1 is used for deep search.}
\label{tab:training_data_stats}
\vspace{-4mm}
\end{wraptable}
\begin{figure*}[t]
\centering
\begin{tcblisting}{
    colback=gray!10,
    colframe=gray!50!black,
    title=Long-Context Document QA System Prompt,
    width=\textwidth,
    listing only,
    boxsep=1mm,
    left=1mm,
    right=1mm,
    top=1mm,
    bottom=1mm,
    listing options={
        basicstyle=\ttfamily\scriptsize,
        breaklines=true,
        breakatwhitespace=false,
        columns=fullflexible,
        keepspaces=true,
        showstringspaces=false
    }
}
You are an AI assistant specialized in long-context document QA. Use the currently available tools to find evidence in the attached document, save useful findings, keep the context compact, and submit the final answer through `finish`.

## Rules
- Use only tools that are currently available in the API payload. If a tool is mentioned here but not currently available, do not call it.
- Keep reasoning before tool calls brief. Put durable facts into `memorize` / `updateMemory` or `note` / `updateNote` instead of long assistant prose.
- When cleaning context, use exact `[msg_id=N]` values from previous messages. Do not target the system message, the original user question, or messages already deleted/truncated/summarized/compressed/folded unless `restoreContext` is available and needed.
- Final answers must be submitted with `finish`; do not answer in plain text outside the tool call.

## Workflow
1. Call `analyzeText` first. If `plan` is available before indexing, write a concise initial strategy. Then call `buildIndex`.
2. Use `searchEngine` for precise keywords, names, dates, titles, rare phrases. If a search returns no useful chunks, try different keywords or queries.
3. Make a `plan` after a successful search.
4. Read evidence with `readChunk` or `readMultiChunks` (respect the tool's chunk-count limit, usually at most 3 ids).
5. After reading, save useful findings with `memorize`, `updateMemory`, `note`, or `updateNote`.
6. After saving notes or memories, clean up bulky context whenever cleanup tools are available:
   - If there are 2 or more uncleaned successful search-result messages, clean search-result tool responses first. Target only the msg_ids of the tool responses of `searchEngine`.
   - If there are 3 or more uncleaned `plan` calls, clean old `plan` assistant messages. Target the assistant msg_ids that invoked `plan`, not the short tool responses.
   - Then prune the current `readChunk` / `readMultiChunks` tool result, and delete the assistant message that invoked `memorize` / `updateMemory` / `note` / `updateNote`.
   - Use `deleteContext` for irrelevant content, `truncateContext` for useful spans, `summarizeContext` for manual summaries, and `compressContext` for automatic compression.
   - Never apply these cleanup tools to the same message twice. If a msg_id has already been deleted, truncated, summarized, or compressed, do not target that msg_id again.
7. Before finishing, if `readNote` or `loadMemory` is available, review the saved notes or memories that are relevant to the answer.
8. Continue searching, reading, memorizing, cleaning, and reviewing until the evidence is sufficient; then call `finish` with a concise answer grounded in the document.
\end{tcblisting}
\caption{System prompt used by \sysname{} for long-context document QA.}
\label{fig:system_prompt}
\end{figure*}

\begin{figure*}[t]
\centering
\begin{tcblisting}{
    colback=gray!10,
    colframe=gray!50!black,
    title=Deep Search System Prompt,
    width=\textwidth,
    listing only,
    boxsep=1mm,
    left=1mm,
    right=1mm,
    top=1mm,
    bottom=1mm,
    listing options={
        basicstyle=\ttfamily\scriptsize,
        breaklines=true,
        breakatwhitespace=false,
        columns=fullflexible,
        keepspaces=true,
        showstringspaces=false
    }
}
You are a web-search QA agent. Answer the user's question with evidence gathered through the currently available tools, keep the conversation compact, and submit the final answer through `finish`.

## Rules
- In search phase, use `search` for batched web queries, `visit` to read promising pages, and `finish` when you have found the answer.
- If `checkBudget` shows that the context length is beyond the threshold, call cleanup tools.
- When cleaning context, use exact `[msg_id=N]` values. Do not target the system message, the original user question, or messages already deleted/truncated/summarized/compressed/folded.
- Final answers must be submitted with `finish`; do not answer in plain text outside the tool call.

## Workflow
1. Use `search` with concise, complementary query strings to explore names, translations, aliases, dates, clue interpretations, and likely sources.
2. Use `visit` on the most promising URLs. Give each visit a precise goal describing what to extract or verify.
3. Iterate search and visit until you have enough evidence. Keep pre-tool reasoning brief and avoid copying large retrieved text into assistant messages.
4. If cleanup tools are available, reduce bulky history before continuing:
   - Prefer large `search` or `visit` results that are no longer needed verbatim.
   - Preserve facts still needed for the final answer: names, aliases, titles, dates, URLs, clue matches, and intermediate conclusions.
   - Use `deleteContext` for irrelevant content, `truncateContext` for useful spans, `summarizeContext` for concise faithful summaries, `compressContext` for automatic compression, and `foldHistory` when available to fold accumulated history.
5. Continue researching if evidence is incomplete; otherwise call `finish` with a concise, evidence-grounded answer.
\end{tcblisting}
\caption{System prompt used by \sysname{} for deep search tasks.}
\label{fig:deep_search_system_prompt}
\end{figure*}

After trajectory generation, we apply a three-stage filtering pipeline.
First, we perform outcome-based filtering with exact match.
For samples that are initially answered incorrectly, we allow two additional retries and retain the trajectory if any retry yields the correct answer.
Second, we use GPT-OSS-120B~\cite{openai2025gptoss120bgptoss20bmodel} for process-based filtering, removing trajectories that exhibit improper context management behavior.
Finally, we discard trajectories whose peak context length exceeds $32$K tokens, ensuring that the retained demonstrations remain compatible with the target context budget.
This procedure starts from $3{,}196$ questions and retains $3{,}114$ trajectories after outcome-based filtering.
Process and peak-token filtering further remove $46$ trajectories, leaving $3{,}068$ qualified trajectories.
These trajectories are then segmented at context editing operations, producing $51{,}469$ SFT trajectory snapshots.

\section{Training Details}
\label{apx:train_details}

We conduct SFT and RL training with the verl library~\cite{sheng2025hybridflow}.
During SFT, we use ZeRO-3 parallelism, a global batch size of $128$, a learning rate of $5 \times 10^{-6}$, a cosine learning-rate scheduler, and a warmup ratio of $0.03$.
For RL, we train for $128$ steps with a rollout batch size of $16$ and set the KL coefficient to $0.001$.
For each query, we first sample $8$ trajectory-level rollouts.
Each complete trajectory is segmented into at most $8$ trajectory snapshots, yielding up to $64$ snapshots from the initial rollouts.
We collect $128$ snapshots per query, and use partial rollout to complete the remaining snapshots.
For the sensitivity score used to select partial-rollout branching points, we set $\alpha=1$ and $\beta=1$.
The maximum input sequence length is $30$K tokens, and the maximum output length is $2$K tokens.
During inference, a temperature of $0.7$ and top-$p$ of $0.8$ are used for our model.
The maximum interaction budget is $60$ turns for long-context QA and $60$ tool calls for deep search following~\citet{wu2025resum}.
The system prompts used by \sysname{} for long-context document QA and deep search are shown in Figures~\ref{fig:system_prompt} and~\ref{fig:deep_search_system_prompt}, respectively.
For LLM-as-a-Judge grading, we use GPT-OSS-120B~\cite{openai2025gptoss120bgptoss20bmodel} with the open-ended judging prompt of StateLM~\cite{liu2026pensieve}, as shown in Figure~\ref{fig:llm_judge_prompt}.

\begin{figure}[t]
\centering
\begin{tcolorbox}[
    colback=gray!10,
    colframe=gray!50!black,
    title=LLM-as-a-Judge Prompt,
    fontupper=\small,
    width=\linewidth
]
Given a problem, its correct answer, and a student's answer below, your task is to review the student's answer and determine if it is correct by comparing it to the correct answer.
If the student's answer is incomplete or ambiguous, assume it is incorrect.

\vspace{10pt}

\#\#\# Problem

\{problem\}

\vspace{10pt}

\#\#\# Answer

\{answer\}

\vspace{10pt}

\#\#\# Student Answer

\{mode\_ans\}

\vspace{10pt}

Please put your final answer (True or False) in \textbackslash \textbackslash boxed\{\}.
Specifically, if the student's answer is correct, the final answer should be \textbackslash \textbackslash boxed\{True\}; otherwise, the final answer should be \textbackslash \textbackslash boxed\{False\}.
\end{tcolorbox}
\caption{Prompt template used by the LLM-as-a-Judge for evaluating open-ended questions, following StateLM~\cite{liu2026pensieve}.}
\label{fig:llm_judge_prompt}
\end{figure}

\section{Baseline Reproduction Details}
\label{apx:baseline_details}

\paragraph{ReSum}
We reproduce the training-free version of ReSum~\cite{wu2025resum} for deep search, using WebSailor-7B and WebExplorer-8B as backbone models.
Following our deep search setting, all runs use a $32$K context window, with $30$K tokens allocated to the input and $2$K tokens reserved for generation; summarization is triggered when the context exceeds $25$K tokens, and the maximum tool call budget is set to $60$.
Since the official ReSum summarization model is not publicly available, we use Qwen3-30B-A3B as the external summarization model.
For the summarizer, we adopt the recommended non-thinking decoding configuration of Qwen3, setting temperature to $0.7$ and top-$p$ to $0.8$.
The context summarization prompt and the summary-conditioned continuation prompt follow the original ReSum prompt.

\paragraph{SUPO}
We reproduce SUPO~\cite{lu2025scaling} using WebSailor-7B and WebExplorer-8B as backbone models.
To isolate the effect of the training data, we train SUPO on the same $1$K OpenSeeker samples as \sysname{}.
Following the design of SUPO, summaries are generated by the policy model itself and optimized jointly with tool use actions.
The summarization threshold is set to $L=0.95\times 30\mathrm{K}=28.5\mathrm{K}$ tokens, where $30$K is our maximum input length.
Following SUPO, we set the maximum number of interaction steps to $H=100$ and the maximum number of summaries to $S=2$, and apply overlong masking to rollouts that fail to produce a final answer before reaching either limit.
The remaining RL hyperparameters are also the same as SUPO: batch size $B=32$, group size $G=8$, learning rate $1\times10^{-6}$ with a constant learning-rate scheduler, and clipping coefficients $\epsilon_{\mathrm{high}}=0.28$ and $\epsilon_{\mathrm{low}}=0.20$.

\section{Additional Token Efficiency Analysis}
\label{apx:token_efficiency}

\begin{figure*}[t]
    \centering
    \begin{subfigure}{0.49\linewidth}
        \centering
        \includegraphics[width=\linewidth]{figures/pdf/token_count/token_per_turn_bars__BrowseComp.pdf}
        \caption{BrowseComp.}
        \label{fig:token_count_appendix_browsecomp}
    \end{subfigure}
    \hfill
    \begin{subfigure}{0.49\linewidth}
        \centering
        \includegraphics[width=\linewidth]{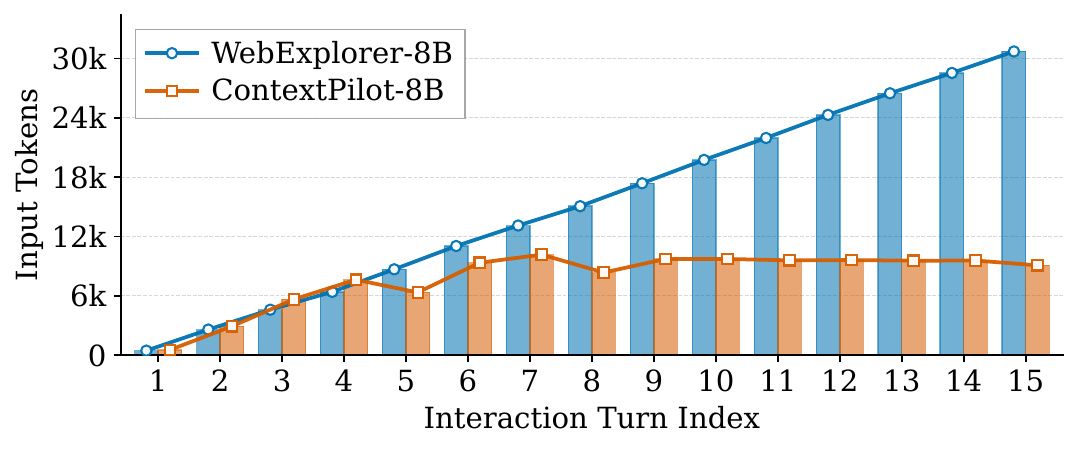}
        \caption{BrowseComp-ZH.}
        \label{fig:token_count_appendix_browsecomp_zh}
    \end{subfigure}
    \caption{Token usage per turn on BrowseComp and BrowseComp-ZH.}
    \label{fig:token_count_appendix}
\end{figure*}

Figure~\ref{fig:token_count_appendix} extends the token-efficiency analysis to both BrowseComp and BrowseComp-ZH.
We consider trajectories with at least $15$ interaction turns and report the average number of input tokens per turn.
Across both benchmarks, the WebExplorer-8B agent accumulates increasingly long contexts as the interaction proceeds.
In contrast, \sysname{} keeps the working context substantially more compact, indicating that its context management tools reduce redundant history without relying on a larger context window.

\begin{table*}[t]
\scriptsize
\centering
\caption{Tool execution failure reason analysis.}
\label{tab:tool_failure_definitions}
\setlength{\tabcolsep}{4pt}
\def\arraystretch{1.12}
\resizebox{\textwidth}{!}{
\begin{tabular}{lp{0.72\textwidth}}
\toprule
\textbf{Tool} & \textbf{Failure Reason} \\
\midrule
\texttt{plan} & The predicted summary is empty or unparseable. \\
\texttt{searchContext} & The search query is empty or malformed, or the model searches before a valid index has been built. \\
\texttt{readChunk} & The requested chunk id is out of range, deleted, or not associated with the current index. \\
\texttt{readMultiChunks} & Any requested chunk id is invalid, duplicated, deleted, or exceeds the allowed batch size. \\
\texttt{note} & The note content is empty, unparseable, or does not specify the value to be stored. \\
\texttt{updateNote} & The target note does not exist, or the update content is empty or unparseable. \\
\texttt{readNote} & The requested note does not exist. \\
\texttt{memorize} & The memory content is empty or unparseable (e.g., inconsistent with the memory storage schema). \\
\texttt{updateMemory} & The target memory item does not exist, or the update content is empty, unparseable, or inconsistent with the memory schema. \\
\texttt{readMemory} & The requested memory item does not exist. \\
\texttt{deleteContext} & The target message index does not exist, has already been deleted, or refers to a protected message (e.g., system prompt or user query). \\
\texttt{summarizeContext} & The target message does not exist, has already been offloaded, or the generated summary is unparseable. \\
\texttt{compressContext} & The target message does not exist, has already been offloaded, or specifies an invalid compression ratio. \\
\texttt{foldHistory} & There is no historical context to fold, or the generated summary and keywords are unparseable. \\
\bottomrule
\end{tabular}
}
\end{table*}

\begin{table*}[t]
\centering
\caption{Complete ablation results of the RL algorithm for Qwen3-8B and Gemma4-E4B-it.
Numbers in parentheses indicate changes relative to the previous line.}
\label{tab:rl_training_ablation_gemma}
\scriptsize
\setlength{\tabcolsep}{3pt}
\def\arraystretch{1.12}
\resizebox{\textwidth}{!}{
\begin{tabular}{llccccc}
\toprule
\textbf{Model} & \textbf{Training method} & \textbf{NovelQA} & \textbf{$\infty$Bench} & \textbf{LongMemEval-S} & \textbf{BrowseComp+} & \textbf{Avg.} \\
\midrule
\multirow{5}{*}{Qwen3-8B}
& SFT & 82.56 & 71.03 & 60.67 & 48.84 & 65.78 \\
& GRPO & 83.53 {\scriptsize \textcolor{green2}{(+0.97)}} & 72.78 {\scriptsize \textcolor{green2}{(+1.75)}} & 60.07 {\scriptsize \textcolor{red}{(-0.60)}} & 50.96 {\scriptsize \textcolor{green2}{(+2.12)}} & 66.84 {\scriptsize \textcolor{green2}{(+1.06)}} \\
& + Entropy-based partial rollout & 82.52 {\scriptsize \textcolor{red}{(-1.01)}} & 73.07 {\scriptsize \textcolor{green2}{(+0.29)}} & 62.13 {\scriptsize \textcolor{green2}{(+2.06)}} & 49.64 {\scriptsize \textcolor{red}{(-1.32)}} & 66.84 {\scriptsize \textcolor{darkgray}{(+0.00)}} \\
& + Context-aware partial rollout & 83.05 {\scriptsize \textcolor{green2}{(+0.53)}} & 73.94 {\scriptsize \textcolor{green2}{(+0.87)}} & 61.40 {\scriptsize \textcolor{red}{(-0.73)}} & 51.08 {\scriptsize \textcolor{green2}{(+1.44)}} & 67.37 {\scriptsize \textcolor{green2}{(+0.53)}} \\
& + Fine-grained credit assignment & \textbf{83.88} {\scriptsize \textcolor{green2}{(+0.83)}} & \textbf{75.25} {\scriptsize \textcolor{green2}{(+1.31)}} & \textbf{64.27} {\scriptsize \textcolor{green2}{(+2.87)}} & \textbf{54.18} {\scriptsize \textcolor{green2}{(+3.10)}} & \textbf{69.40} {\scriptsize \textcolor{green2}{(+2.03)}} \\
\midrule
\multirow{5}{*}{Gemma4-E4B-it}
& SFT & 66.80 & 55.02 & 55.07 & 42.05 & 54.74 \\
& GRPO & 71.12 {\scriptsize \textcolor{green2}{(+4.32)}} & 57.06 {\scriptsize \textcolor{green2}{(+2.04)}} & 57.00 {\scriptsize \textcolor{green2}{(+1.93)}} & 43.41 {\scriptsize \textcolor{green2}{(+1.36)}} & 57.15 {\scriptsize \textcolor{green2}{(+2.41)}} \\
& + Entropy-based partial rollout & 70.81 {\scriptsize \textcolor{red}{(-0.31)}} & 58.22 {\scriptsize \textcolor{green2}{(+1.16)}} & 60.67 {\scriptsize \textcolor{green2}{(+3.67)}} & 46.27 {\scriptsize \textcolor{green2}{(+2.86)}} & 58.99 {\scriptsize \textcolor{green2}{(+1.85)}} \\
& + Context-aware partial rollout & 71.24 {\scriptsize \textcolor{green2}{(+0.43)}} & 59.68 {\scriptsize \textcolor{green2}{(+1.46)}} & 60.33 {\scriptsize \textcolor{red}{(-0.34)}} & 46.14 {\scriptsize \textcolor{red}{(-0.13)}} & 59.35 {\scriptsize \textcolor{green2}{(+0.36)}} \\
& + Fine-grained credit assignment & \textbf{72.92} {\scriptsize \textcolor{green2}{(+1.68)}} & \textbf{60.99} {\scriptsize \textcolor{green2}{(+1.31)}} & \textbf{62.47} {\scriptsize \textcolor{green2}{(+2.14)}} & \textbf{47.47} {\scriptsize \textcolor{green2}{(+1.33)}} & \textbf{60.96} {\scriptsize \textcolor{green2}{(+1.62)}} \\
\bottomrule
\end{tabular}
}
\end{table*}

\section{Tool Failure Definitions}
\label{apx:tool_failure_definitions}

We consider a tool invocation failed if its execution triggers an environment-side error.
Concretely, this includes malformed calls, invalid arguments, and violations of tool-specific preconditions.
For retrieval tools, returning no matched evidence is not counted as a failure if the query and index are valid.
Table~\ref{tab:tool_failure_definitions} summarizes the specific cases that may lead to tool execution errors in our analysis.
Figure~\ref{fig:tool_error} reports the execution failure rate for different tool categories on $\infty$Bench and NovelQA.
Across both benchmarks, memory-related and context offloading tools have higher failure rates at the early stage of RL training, indicating that the SFT model can invoke these tools but has not yet fully learned their correct usage conditions.
These failure rates decrease substantially during RL training, showing that RL improves the reliability of context management operations while reshaping the model's tool use strategy.

\section{Additional RL Training Ablation}
\label{apx:rl_training_ablation}

Table~\ref{tab:rl_training_ablation_gemma} reports the complete RL training ablation results on both Qwen3-8B and Gemma4-E4B-it.
The results show a consistent trend across the two models: context-aware partial rollout improves over entropy-only branching, and fine-grained credit assignment further strengthens the final average performance.

\section{Use of Scientific Artifacts}
\label{apx:scientific_artifacts}

We cite the original creators of scientific artifacts wherever they are introduced.
For models, Qwen3 and Gemma4 are released under the Apache License 2.0.
For datasets, OpenSeeker-V1 is released under the MIT License, while the other datasets used in this work are released under the Apache License 2.0.
We use all artifacts only for research purposes and in a manner consistent with their intended use in benchmarking, model training, and evaluation.
We do not redistribute the original datasets or model weights as part of this work.
The artifacts cover two primary task domains: long-context QA and deep search. 
Language includes English and Chinese.
We do not collect new data from human subjects.
Because the experiments rely on existing public benchmarks and datasets, we do not perform additional manual anonymization or offensive-content filtering.
Relevant data statistics, including the number of questions, retained trajectories, and trajectory snapshots used for training, are reported in Table~\ref{tab:training_data_stats}.

\section{Use of AI Assistants}
\label{apx:ai_assistants}

We used Codex to assist with code modification, and ChatGPT for writing polishing.
All assisted outputs were manually reviewed, verified, and finalized by the authors.

\end{document}